\documentclass{ecai} 

\usepackage{latexsym}
\usepackage{amssymb}
\usepackage{amsmath}
\usepackage{amsthm}
\usepackage{booktabs}
\usepackage{enumitem}
\usepackage{graphicx}
\usepackage{color}
\usepackage{adjustbox}
\usepackage{textcomp}
\usepackage{xcolor}
\usepackage{enumitem}
\usepackage{times}
\usepackage{soul}
\usepackage{url}
\usepackage[hidelinks]{hyperref}
\usepackage[utf8]{inputenc}
\usepackage[small]{caption}
\usepackage{graphicx}
\usepackage{amsmath}
\usepackage{amsthm}
\usepackage{booktabs}
\usepackage{algorithm}
\usepackage{algorithmic}
\usepackage{multirow}
\usepackage{multicol}
\usepackage{amssymb}
\usepackage{graphicx}  
\usepackage{float}  
\usepackage{subfigure}  
\usepackage{indentfirst}
\usepackage[switch]{lineno}

\newcommand{\BibTeX}{B\kern-.05em{\sc i\kern-.025em b}\kern-.08em\TeX}

\begin{document}


\begin{frontmatter}




\title{PMR: Physical Model-Driven Multi-Stage Restoration of Turbulent Dynamic Videos}


\author[A]{\fnms{Tao}~\snm{Wu}\footnote{Equal contribution.}}
\author[A]{\fnms{Jingyuan}~\snm{Ye}\footnotemark}
\author[A]{\fnms{Cheng}~\snm{Zhou}} 
\author[A]{\fnms{Wenlong}~\snm{Chen}}
\author[B]{\fnms{Zheng}~\snm{Liu}} 

\author[B]{\fnms{Huiming}~\snm{Zheng}} 
\author[C]{\fnms{Wei}~\snm{Liu}} 
\author[A]{\fnms{Ying}~\snm{Fu}\thanks{Corresponding author. Email: fuying@cuit.edu.cn.}}

\address[A]{Chengdu University of Information Technology}
\address[B]{National Innovation Center For UHD Video Technology}
\address[C]{School of Automation and Intelligent Sensing \& Institute of Image Processing and Pattern Recognition \& Institute of Medical Robotics, Shanghai Jiao Tong University}


\begin{abstract}
Geometric distortions and blurring caused by atmospheric turbulence degrade the quality of long-range dynamic scene videos. Existing methods struggle with restoring edge details and eliminating mixed distortions, especially under conditions of strong turbulence and complex dynamics. To address these challenges, we introduce a Dynamic Efficiency Index ($DEI$), which combines turbulence intensity, optical flow, and proportions of dynamic regions to accurately quantify video dynamic intensity under varying turbulence conditions and provide a high-dynamic turbulence training dataset. Additionally, we propose a Physical Model-Driven Multi-Stage Video Restoration ($PMR$) framework that consists of three stages: \textbf{de-tilting} for geometric stabilization, \textbf{motion segmentation enhancement} for dynamic region refinement, and \textbf{de-blurring} for quality restoration. $PMR$ employs lightweight backbones and stage-wise joint training to ensure both efficiency and high restoration quality. Experimental results demonstrate that the proposed method effectively suppresses motion trailing artifacts, restores edge details and exhibits strong generalization capability, especially in real-world scenarios characterized by high-turbulence and complex dynamics. We will make the code and datasets openly available.

\end{abstract}


\end{frontmatter}


\section{Introduction}
Atmospheric turbulence, an optical phenomenon caused by spatiotemporal variations in the air's refractive index, is influenced by factors such as temperature, wind speed, air pressure, and the distance between the target and the imaging device. In complex dynamic conditions, such as extreme temperature gradients \cite{10.1117/12.956537} or rapid target movements \cite{XU201987}, turbulence severely disrupts the imaging over long horizontal and slanted paths. This disruption leads to pixel displacements due to geometric distortions and localized non-uniform blurring, significantly degrading image and video quality. Recent image and video enhancement techniques aim to address turbulence-induced distortions. However, these methods struggle with effective and synchronized recovery in turbulent dynamic scenes, owing to the complexity of the distortions. \textbf{Current research} faces several limitations: \(\mathbf{1)}\) Difficulty effectively distinguishing turbulence interference from object motion when there are mixed distortions. \(\mathbf{2)}\) The physical model analysis of turbulent dynamic videos is insufficient\par  

\begin{figure}[h]
  \centering
  \raisebox{-0.5\height}{%
    \resizebox{0.5\textwidth}{4.8cm}{%
      \includegraphics{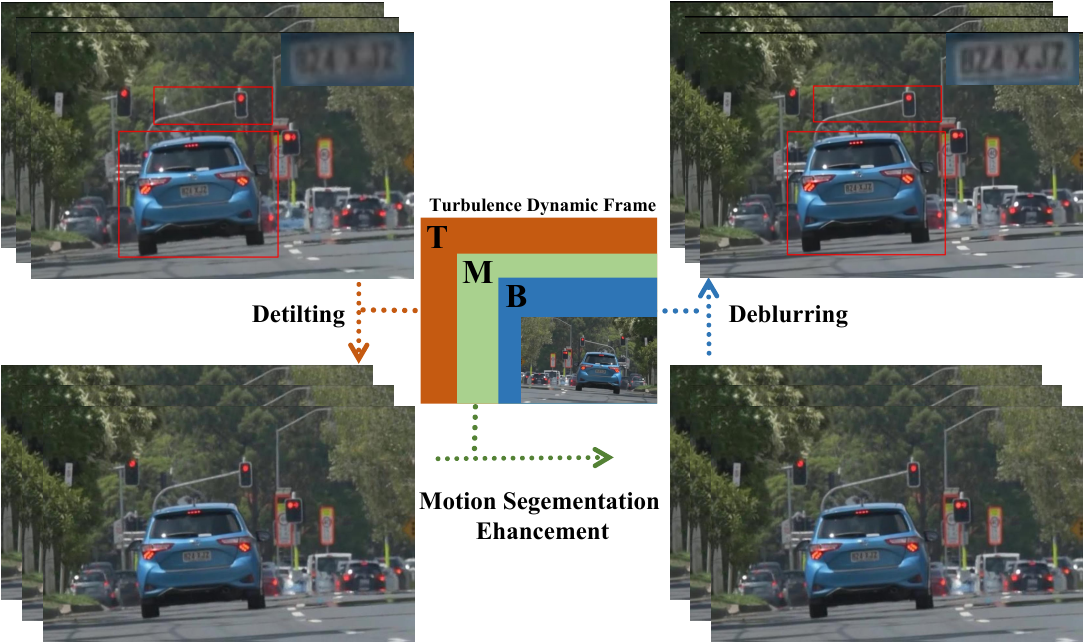}%
    }%
  }
  \caption{Guided by the physical model of turbulent dynamic video frames, our proposed multi-stage restoration on real-world datasets \citep{Jin2021NeutralizingTI} effectively recovers edge details of various objects and enhances license plate readability.}
  \vspace{12pt} 
  \label{fig1}
\end{figure}

\noindent to fully understand mixed distortions, limiting effective restoration. \(\mathbf{3)}\) Existing methods are mainly designed for static scenes or scenarios with minor motion, and lack practical and effective approaches for complex distortion scenarios with large motion. To address these challenges, this paper makes the following major contributions:

\begin{enumerate}[label=\arabic{enumi})]
    \item The Dynamic Efficiency Index ($DEI$) based on turbulence intensity is designed to accurately assess the impact of turbulence on video quality in dynamic scenes. This index combines optical flow correction with dynamic region proportion, providing a basis for dynamic analysis under varying turbulence conditions. 
    \item We propose a Physics-Based Multi-Stage Restoration ($PMR$) framework for staged restoration of mixed distortions in turbulent dynamic videos, organized into three main stages: \textbf{De-tilting}. Which corrects geometric distortions and pixel displacements by integrating multi-scale optical flow information. \textbf{Motion Segmentation Enhancement}, which segments the foreground and background, enhancing the target area to highlight prominent features. \textbf{De-blurring}, which combines spatiotemporal dynamic characteristics to capture local details and correct blurred regions, producing high-quality images. 
    \item In the De-tilting and De-blurring stages of the PMR framework, we design two lightweight networks respectively, aiming to balance computational resource consumption and processing speed with restoration performance.
\end{enumerate}





\section{Related Work}
\subsection{ Physical Modeling of Turblence}
Since the foundational work of \cite{Kolmogorov1991TheLS} and \cite{Fried_1978}, the physical properties of turbulence have been studied systematically. Extensive parametric modeling efforts simulate the refractive phenomena of turbulence \cite{10.1145/1276377.1276451} \cite{Hardie2017SimulationOA}, with turbulence models based on wave propagation equations widely used in related research \cite{book}. Currently, the Zernike coefficient decomposition method in phase space is the mainstream for turbulence simulation \cite{Roggemann:95} \cite{9711075}. This approach decomposes complex phase fluctuations into components, with the first two Zernike coefficients describing the pixel displacement (tilt) and the higher-order coefficients characterizing aberrations (blur). Prioritizing tilt correction followed by blur removal significantly improves modeling accuracy and effectiveness \cite{9864020}. \par

Based on this theory, we model turbulent dynamic videos from a video restoration perspective, as shown in Figure \ref{fig1}. Turbulence interference is decomposed into a combination of tilt distortion (\(T\)) and blur distortion (\(B\)), while motion artifacts (\(M\)) in dynamic scenes are treated as distortions over a static background. This relationship is formalized as follows:

\begin{equation}
\begin{aligned}
\mathrm{I}_{\mathrm{i}}(\mathrm{x}, \mathrm{t}) & =\mathrm{M}\left\{\mathrm{T}^{\circ} \mathrm{B}\right\}\left(\mathrm{J}_{\mathrm{i}}(\mathrm{x}, \mathrm{t})\right) \\
& =\mathrm{T}\left(\mathrm{M}\left(\mathrm{B}\left(\mathrm{J}_{\mathrm{i}}(\mathrm{x}, \mathrm{t})\right)\right)\right)
\end{aligned}
\tag{1}\label{E1}
\end{equation}

\noindent where \(\mathrm{J}_{\mathrm{i}}(\mathrm{x}, \mathrm{t})\) denotes a clear frame, \(x \in R ^ { 2 }\) represents 2D spatial coordinates, $t$ corresponds to the temporal dimension, and $ \mathrm{T}^{\circ} \mathrm{B}$ represents distortion of the turbulence. Building upon this model, we propose a multi-stage restoration framework that progressively eliminates $T$, $M$, and $B$, recovering high-quality dynamic frames.

\subsection{Single-Stage Turbulence Restoration Methods}
Recent deep-learning methods treat turbulence disturbances as a single type of distortion, performing restoration in an end-to-end manner. For example, CNN-based methods \cite{9762752}\cite{ANANTRASIRICHAI202369} effectively learn turbulence characteristics, with multi-frame processing outperforming single-frame approaches \cite{Zhang_Zuo_Chen_Meng_Zhang_2017}. However, due to static filter weights and limited receptive fields, CNNs struggle with spatial dynamics. Mao et al. proposed TurbNet \cite{10.1007/978-3-031-19800-7_25}, integrated physics-inspired degradation and reconstruction modules, and replaced transformers with convolutions for enhanced modeling. Wu et al. introduced a semi-supervised self-attention model using the Mean Teacher method for leveraging unlabeled data \cite{article}. The TSR-WGAN model \cite{Jin2021NeutralizingTI} combines spatio-temporal information, treating turbulence videos as 3D tensors to learn residuals. López-Tapia et al. integrated an RNN into the GAN generator to predict optical flow fields for dynamic objects \cite{10222374}, achieving better performance than AT-Net \cite{9506614}. Jaiswal al et. have utilised an attention based network (Swin) combined with a diffusion for reportedly state-of-the-art results.\cite{Jaiswal2023PhysicsDrivenTI}.

\subsection{Multi-Stage Turbulence Restoration Methods}
Compared to single-stage methods, multi-stage strategies restore edge details more accurately by decomposing mixed distortions, improving visual quality. The TMT method \cite{TMT} employs CNNs for geometric correction and transformers for detail reconstruction. AT-Net \cite{9506614} uses a dual UNet architecture for geometric correction and blur removal. Shimizu et al. proposed a three-stage approach \cite{4587525} that combined frame averaging, B-spline registration, and multi-frame reconstruction. In diffusion-based research, Chung et al. introduced a method \cite{Chung2022ParallelDM} using diffusion models as a prior for blind restoration, similar to the two-stage correction in TMT. Jiang et al. extended implicit neural representation grid deformation \cite{10208382} to incorporate realistic tilt and blur models \cite{9710477}.

\section{Method}
This section is divided into two parts: we first introduce the dynamic efficiency index to quantify dynamic effects in turbulent videos; we then present the multi-stage restoration method and core model design. \par

\subsection{Dynamic Efficiency Index (DEI)}
Atmospheric turbulence induces pixel shifts in video frames, creating pixel dancing. Existing methods for quantifying dynamic intensity in turbulent videos struggle to differentiate between turbulence-induced shifts and object motion. To address this, we introduce the $DEI$, which combines turbulence intensity estimation \(C_n^2\), dynamic optical flow maps (\(DyOF_{i}\)), and Dynamic Region Proportion ($DPR$) to correct pixel displacements, mitigating turbulence interference and enabling precise dynamic quantification.
The calculation proceeds as follows: \textbf{First}, we compute the turbulence intensity \(C_n^2\) \cite{Saha:22}\cite{Zamek2006TurbulenceSE}:

\begin{equation}
\begin{aligned}
C_n^2 = \frac{PFOV^2 \times D^{\frac{1}{3}}}{L \times P} \times \frac{Var(V)}{Grad^n(V)},
\end{aligned}
\tag{2}\label{E2}
\end{equation}

\noindent where \(PFOV\) is the pixel field of view, \(D\) is the lens diameter, \(L\) is the target distance, \(P\) is the turbulence constant and \(V\) is the sequence of images. The work in \cite{Zamek2006TurbulenceSE} shows that as \(C_n^2\) increases, the turbulence effects intensify, and lower values indicate weaker turbulence interference. \textbf{Next}, we compute the optical flow between the current frame \(i\) and its adaptive selection of neighboring \(N\) frames using the $RAFT$ model \cite{10.1007/978-3-030-58536-5_24}. We then average the resulting \(N-1\) optical flow maps to smooth inter-frame variations and reduce noise impact. Finally, adaptive adjustments based on \(C_n^2\) are performed as follows:

\begin{equation}
OF_i = \frac{\sum_{j=i}^{N} RAFT(F_j, F_{j+1})}{N-1} \times \frac{1}{1 + C_n^2}.
\tag{3}\label{E3}
\end{equation}

After normalizing the magnitude of the adjustment optical flow, we compute its average distance ($d_{avg}$) from the predefined threshold of \(0.5\) and apply binarization, with values near \(1\) indicating dynamic pixels for dynamic region segmentation. The \(DPR\) is then calculated as:

\begin{equation}
\resizebox{0.42\textwidth}{!}{$DyOF_i = Max(d_{avg}(Norm(OF_i), 0.5)), \quad DPR = \frac{\sum_{j=1}^N DyOF_j}{\sum_{j=1}^N OF_j}$}
\tag{4}\label{E4}
\end{equation}

The experimental results indicate that dynamic effects are most pronounced when \(DPR \in [0.05, 0.5]\) while avoiding excessive \(DPR\) to prevent redundancy in information or visual interference. To further quantify the visual dynamic effect, we define the $DPR$ influence coefficient \(C\) as:

\begin{equation}
C = 
\begin{cases} 
2, & 0 \leq \textit{DPR} < 0.05 \\
1, & 0.05 \leq \textit{DPR} < 0.5 \\
0.5, & 0.5 \leq \textit{DPR} < 0.7 \\
0.1, & 0.7 \leq \textit{DPR} < 1.
\end{cases}
\tag{5}
\end{equation}

\begin{figure*}[h]
  \centering
  \raisebox{-0.5\height}{%
    \resizebox{1\textwidth}{7.5cm}{%
      \includegraphics{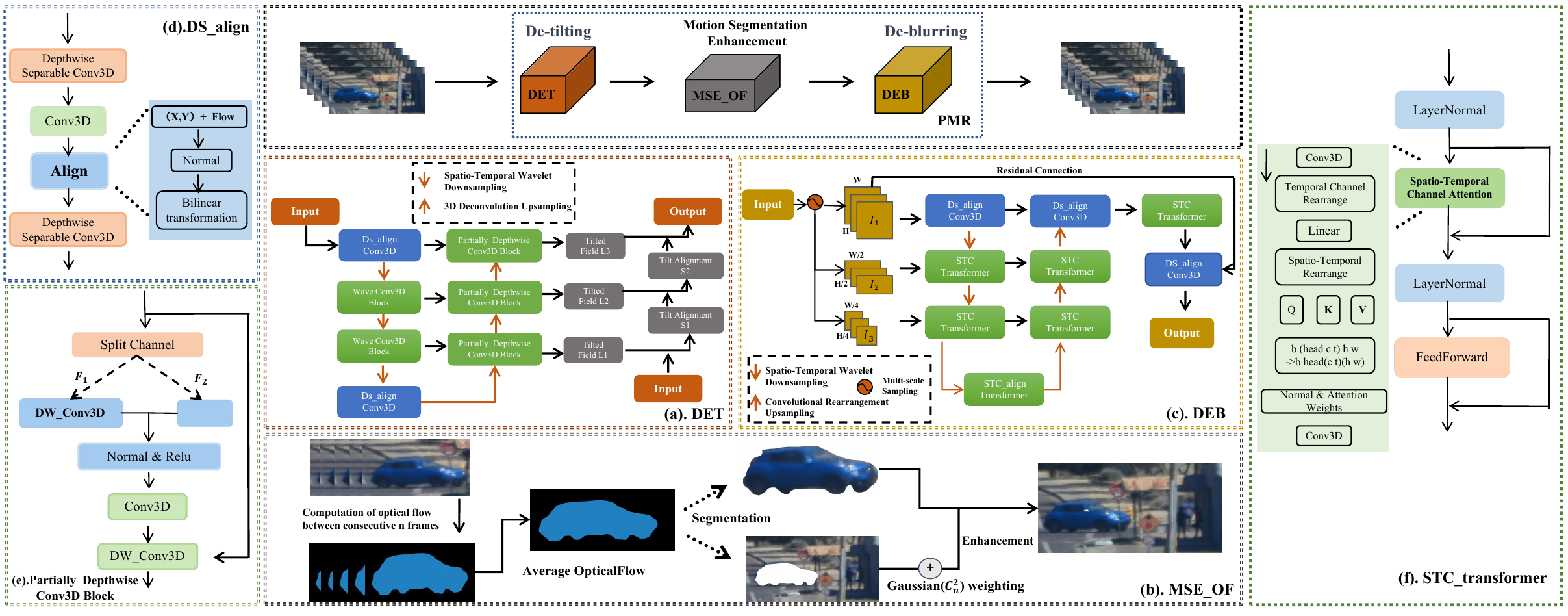}%
    }%
  }
  \caption{Illustation of the turbulence dynamic video restoration process. Based on the physical model described in Equation (\ref{E1}), a multi-stage restoration strategy is employed, consisting of de-tilting, motion segmentation enhancement, and de-blurring.}
  \label{fig2}
\end{figure*}

\textbf{Finally}, by using \(DyOF_{i}\) and \(DPR\), we quantify the dynamic intensity of turbulent videos. Due to turbulence-induced pixel displacements, the directly calculated \(DEI\) may overestimate the motion intensity. To objectively quantify the dynamic intensity, we apply dynamic normalization by incorporating \(C_n^2\) and the turbulence constant \(\gamma\), and the proposed $DEI$ is finally defined as : 

\begin{equation}
DEI = \frac{C}{\gamma} \times \frac{\sum_{i=1}^N DyOF_i}{1 + C_n^2}.
\tag{6}\label{E6}
\end{equation}

\noindent A higher $DEI$ indicates a larger dynamic intensity in the turbulent video, while a lower $DEI$ indicates a smaller intensity.

\subsection{Multi-Stage Restoration Framework}
We propose a Physical Model-Guided Multi-Stage Video Restoration ($PMR$) framework to address spatio-temporal distortions caused by atmospheric turbulence and mixed distortions in dynamic scenes, shown in Figure \ref{fig2}. The framework, based on the physical model defined in Equation (\ref{E1}), divides the restoration of turbulent dynamic videos into three stages: de-tilting, motion segmentation enhancement, and de-blurring. These stages correspond to the following core challenges: \(\mathbf{1)}\) The random temporal displacements of video frame pixels leading to wave-like distortions and jittering phenomena; \(\mathbf{2)}\) The high-motion intensity objects exhibiting artifacts at their edges due to turbulence interference; \(\mathbf{3)}\) The local regions suffering from spatially non-uniform blurring caused by higher-order aberrations.

\begin{figure}[h]
  \centering
  \raisebox{-0.5\height}{%
    \resizebox{0.5\textwidth}{4cm}{%
      \includegraphics{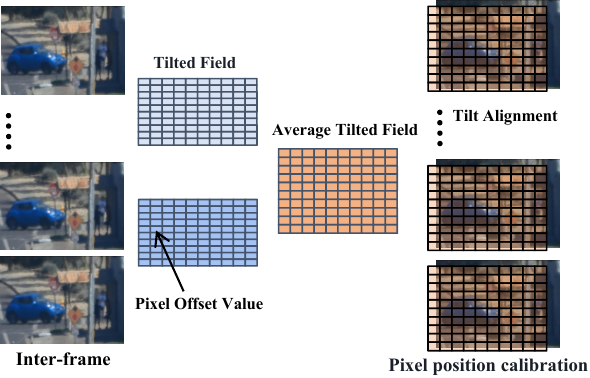}%
    }%
  }
  \caption{Illustrates the correction of temporally random pixel fluctuations using the average tilt field.}
  \label{fig3}
\end{figure}

\subsubsection{De-tilting}
Atmospheric turbulence causes random shifts in the pixel positions of video frames over time, visually appearing as wavy distortions. Based on the tilt displacement following a zero-mean Gaussian process \cite{Fried_1978} \cite{9710477}, we propose a lightweight de-tilting model ($DET$). As shown in Figure \ref{fig2}.(a), the $DET$ model adopts an encoder-decoder architecture with a depth of 4. The first and fourth layers of the decoder extract features and correct pixels using the $DS\_align$ module, while the other encoder layers use the $Wave\_conv3D$ module, which integrates spatiotemporal wavelet downsampling to capture frequency domain and spatiotemporal channel features. The decoder layers utilize the $Part\_conv3D$ module to fuse the upsampled features with skip connection information. In the last three layers of the decoder, multi-scale continuous frame tilt fields are generated to describe pixel displacements. Due to the zero-mean Gaussian nature of tilt distortion, a more table tilt field is obtained by averaging over the temporal domain. Pixel correction is then performed through interpolation mapping, ultimately generating the de-tilted image as shown in Figure \ref{fig3}. The key components of the $DET$ model are as follows:\par

\textbf{DS\_align} module first uses depthwise separable convolutions to extract image features. Then, 3D convolutions map the features to the 2D spatial domain to perform multi-frame pixel correction along the horizontal and vertical axes. Finally, depthwise separable convolutions are applied again to enhance the representation of detailed features, as shown in Figure \ref{fig2}.(d).\par

\textbf{Wave\_conv3D} module integrates depthwise separable convolution with spatiotemporal wavelet downsampling. Each frame is independently decomposed using wavelets to extract high-frequency edge details and low-frequency global structural information. The frames, processed by depthwise separable convolution, are then stacked along the temporal dimension, preserving the wavelet features of each frame. This creates a tensor that retains spatiotemporal joint features for subsequent feature learning. This operation is defined as:

\begin{equation}
    \begin{aligned}
    I(x,t)^{\prime}=append(t)\{Harr[I(x,t)]\}.
    \end{aligned}
    \tag{7}
\end{equation}

\noindent where \(Harr\) denotes the wavelet decomposition, and \(t\) represents the time dimension.

\begin{table*}[h]
    \caption{Quantitative evaluation of the results produced by different methods, the best results are in bold and the second best results are underlined. All the methods are tested on the Normal datasets \cite{TMT} and High DEI turbulent dynamic dataset. * denotes multi-stage turbulence restoration methods.}
    \centering
    \setlength{\tabcolsep}{5pt} 
    \renewcommand\arraystretch{1.2} 
    \small 
    \begin{tabular}{c | c | c c c | c c c}
    \toprule
    \multirow{2}*{\textbf{Methods}} & \multirow{2}*{\textbf{Reference}} & \multicolumn{3}{c}{\textbf{Normal}} & \multicolumn{3}{c}{\textbf{High DEI}}\\ \cline{3-8}
    & & PSNR$\uparrow$ & SSIM$\uparrow$ & LPIPS$\downarrow$ & PSNR$\uparrow$ & SSIM$\uparrow$ & LPIPS$\downarrow$ \\ \hline
    TSRWGAN\cite{Jin2021NeutralizingTI} &NMI-2021 & 26.235 & 0.781 & 0.268 & 26.395 & 0.788 & 0.259 \\
    Restormer\cite{RESTORMER} &CVPR-2022 & 23.781 & 0.712 & 0.351 & 23.832 & 0.714 & 0.348 \\
    TurbNet\cite{10.1007/978-3-031-19800-7_25} &ECCV-2022 & 24.152 & 0.713 & 0.421 & 24.352 & 0.721 & 0.409 \\
    Uformer\cite{uformer} &CVPR-2022 & 23.652 & 0.703 & 0.391 & 23.812 & 0.716 & 0.382 \\
    Histoformer\cite{histoformer} &ECCV-2024 & 25.781 & 0.752 & 0.311 & 25.972 & 0.754 & 0.305 \\
    VRT\cite{10462902} &TPAMI-2024 & 27.282 & 0.812 & \underline{0.252} & 27.530 & 0.821 & 0.250 \\ 
    $*$TurbSR\cite{Saha2024TurbSegResAS} & CVPR-2024 & 26.682 & 0.806 & 0.258 & 27.141 & 0.816 & 0.255 \\
    $*$Deturb\cite{DETURB} & ACCV-2024 & 26.532 & 0.798 & 0.268 & 27.052 & 0.805 & 0.256 \\
    $*$TMT\cite{TMT} & TCI-2024 & \underline{27.593} & \underline{0.820} & \textbf{0.249} & 27.853 & \underline{0.832} & 0.246 \\ \hline
    $*$PMR\_R & Ours & 27.375 & 0.818 & 0.261 & 27.638 & 0.827 & 0.250 \\
    $*$PMR & Ours & 27.31 & 0.815 & \underline{0.252} & \underline{27.862} & \underline{0.832} & \underline{0.245} \\
    $*$PMR\_T & Ours & \textbf{27.665} & \textbf{0.823} & \textbf{0.249} & \textbf{28.223} & \textbf{0.845} & \textbf{0.241} \\ \bottomrule
    \end{tabular}
    \label{tab:1}
\end{table*}

\textbf{Part\_conv3D} Depth convolution is first applied to specific channels, followed by concatenation with the original unprocessed features along the channel dimension. Finally, depthwise separable convolution is used for the overall processing of characteristics, as shown in Figure \ref{fig2}.(e).\par

\subsubsection{Motion Segmentation Enhancement}
In turbulence dynamic video restoration, the de-tilting stage corrects pixel offsets by superimposing multi-scale average tilt fields. However, since the tilt field is applied uniformly across the image, unnecessary adjustments may occur at object motion boundaries, leading to blurring around the boundaries between foreground and background. To address this problem, we propose a method for motion segmentation enhancement based on optical flow ($MSE\_OF$)  as shown in Figure \ref{fig2}.(b).\par

After segmentation of the foreground and background, dynamic regions retain original details to ensure motion authenticity, we apply Gaussian weighting adaptively based on \(C_n^2\) in the static regions. To smooth the background and eliminate turbulence interference, effectively reducing ghosting and artifacts.\par

The key steps are as follows: \textbf{First}, calculate the optical flow maps between consecutive frames as shown in Equation (\ref{E3}) and determine the best optical flow map by calculating the optical flow deviation $(OFD)$ between frames, defined as:

\begin{equation}
\resizebox{0.42\textwidth}{!}{$
\begin{aligned}
OFD &= \text{Min}\left\{ \frac{\max[OF_i]}{2} - \text{Mean}\left( \left| \frac{\max[OF_i]}{2} - OF_i \right| \right) \right\},
\end{aligned}
$}
\tag{8}
\end{equation}

\noindent where $OF_{i}$ denotes the optical flow map of a single frame, \(\max(OF_{i})\) is the global maximum pixel value and $Mean$ represents the average deviation. A lower value \(OFD\) corresponds to a more uniform optical flow magnitude distribution, which produces the optimal optical flow mask.\par

\textbf{Next}, based on the optimal mask, the dynamic and static regions are segmented using Equation (\ref{E4}). Gaussian weights for background regions are computed via \(C_n^2\) and applied at the pixel level with the current frame to achieve Gaussian weighting, thus stabilizing the background and mitigating turbulence effects. The calculation of current frame $i$ Gaussian weights is as follows: 

\begin{equation}
    \begin{aligned}
    W_i = \exp\left(-\frac{(N - i)^2}{2 (C_n^2)^2}\right).
    \end{aligned}
    \tag{9}
\end{equation}

\subsubsection{De-blurring}
After the de-tilting and segmentation enhancement stages, pixel shifts and motion artifacts are largely addressed. However, localized non-uniform blurring caused by higher-order aberrations persists. To further improve image quality, we design a lightweight hybrid $DEB$ model that integrates CNN with a spatio-temporal channel attention mechanism. As shown in Table \ref{tab：2}, the $DS\_align$ module is employed in the first layer of $DEB$ to perform feature extraction and pixel alignment via convolution, effectively reducing the computational burden of transformer blocks on high-resolution inputs. In addition, the number of $STC\_transformer$ blocks is controlled in the subsequent three layers: more blocks are allocated at lower-resolution stages to enhance spatiotemporal-channel attention modeling for better recovery of locally blurred details, while fewer blocks are used at higher-resolution stages to avoid unnecessary receptive field expansion and parameter redundancy.\par

\begin{table}[H]
    \caption{Structural Design of the Lightweight Hybrid Model DEB}
    \centering
    \setlength{\tabcolsep}{3pt} 
    \renewcommand\arraystretch{1.2} 
    \small 
    \begin{tabular}{l|l|l}
        \hline
        Aspects  & Encoder Level [1,2,3,4] & Dncoder Level[3,2,1]  \\
        \hline
        Type of module  & [Ds$\_$align, STC, STC, STC] & [STC, STC, Ds$\_$align]   \\
        Num of module   & [1, 1, 2, 4]     & [2, 1, 1]   \\
        \hline
    \end{tabular}
    \label{tab：2}
\end{table}

As shown in Figure \ref{fig2}.(c), the model $DEB$ first divides the input image into three hierarchical levels of different resolutions for feature extraction in the encoder layers. Through the $DS\_align$ and $STC\_transformer$ modules, the model facilitates cross-resolution contextual interactions, capturing richer and more accurate spatio-temporal features. In the decoding phase, the model gradually restores image details through multi-scale feature fusion. In the final output stage, the $STC\_transformer$ module enhances both global and local characteristics, aligning the enhanced result with the original image via the $DS\_align$ module to produce a clear image. 
\textbf{The key design of the STC\_transformer module is as follows}: It starts with a 3D convolution to perform a grouped channel convolution, fusing the temporal frame and channel features, and uses linear layers to model complex spatio-temporal relationships between frames, as shown in Figure \ref{fig2}.f. The extracted feature maps are divided into query (Q), key (K), and value (V) components along the channel dimension. Q and K are further split into multiple subspaces and processed using a multi-head attention mechanism to compute weights in parallel, effectively capturing key features and enhancing the model's representation capacity. By calculating attention features

\begin{figure*}[h]
  \centering
  \raisebox{-0.5\height}{%
    \resizebox{1\textwidth}{4.4cm}{%
      \includegraphics{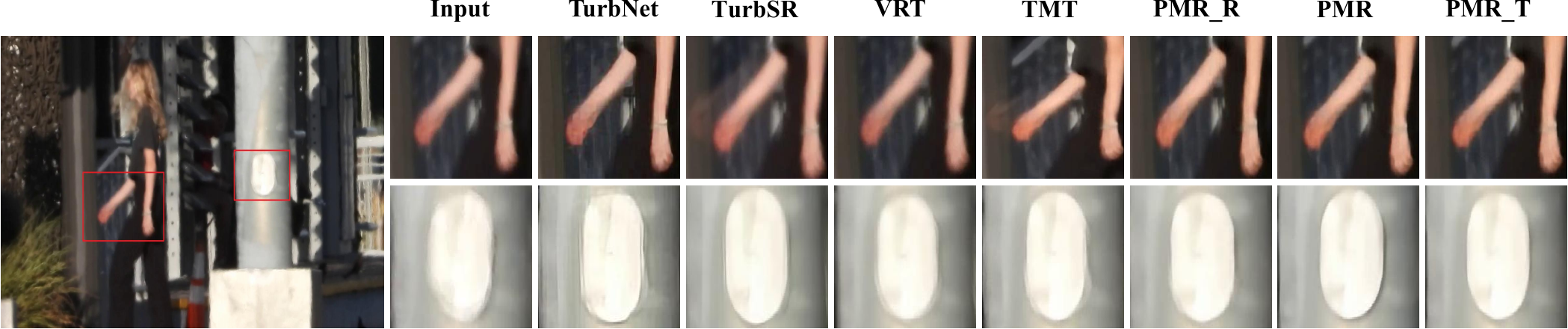}%
    }%
  }
  \caption{Comparison of recovery results for dynamic and static regions in real 2k turbulent videos under close-range shooting.}
  \label{fig4}
\end{figure*}

\begin{figure*}[h]
  \centering
  \raisebox{-0.5\height}{%
    \resizebox{1\textwidth}{4.4cm}{%
      \includegraphics{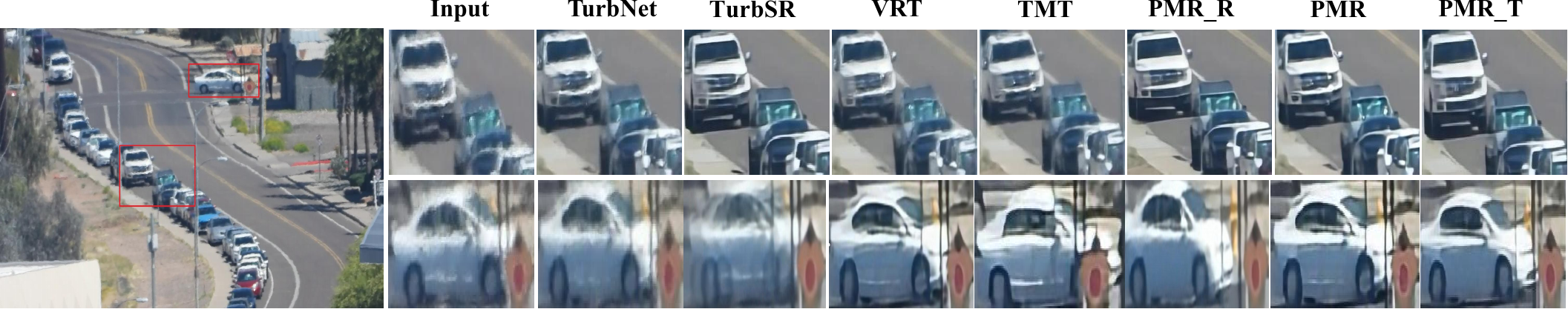}%
    }%
  }
  \caption{Comparison of recovery results for dynamic and static regions in real 2k turbulent videos under long-distance shooting.}
  \label{fig5}
\end{figure*}

\noindent in the temporal dimension's high-frequency domain, the module models dynamic changes between channels. While reducing parameter count and computational complexity, it ensures comprehensive spatio-temporal information fusion.


\subsection{Analysis of the PMR Method} 
The $PMR$ method we propose employs two lightweight models to balance real-time performance and restoration effectiveness in a multi-stage process. To validate the effectiveness of $PMR$ in restoring turbulent dynamic scenes and its generalization ability in real-world scenarios, we replace the lightweight $DEB$ model with two transformer models $Restormer$ \cite{RESTORMER} and $TMT$ \cite{TMT} which have shown promising performance in deblurring tasks. Two variants, $PMR\_R$, and $PMR\_T$, are constructed and validated in subsequent experiments.

\section{Experiments}

\subsection{Implementation Details} 
We compare $PMR$, $PMR\_T$, and $PMR\_R$ with state-of-the-art turbulence restoration methods, including $TSRWGAN$ \cite{Jin2021NeutralizingTI}, $TMT$ \cite{TMT}, $VRT$ \cite{10462902}, $Restormer$ \cite{RESTORMER}, $Uformer$\cite{uformer}, $ATNet$ \cite{9506614}, $Deturb$\cite{DETURB}, $Histoformer$\cite{histoformer}, $TurbNet$ \cite{10.1007/978-3-031-19800-7_25}, and $TurbSR$ \cite{Saha2024TurbSegResAS}. To enhance model learning for turbulent dynamic videos, we classify the synthetic dataset \cite{TMT} based on the $DEI$ metric from Equation (\ref{E6}). Among the 3,500 synthetic turbulent dynamic videos covering 32 different types of motion, we selected 1,568 videos as the high subset $DEI$ for feature-enhanced training, using $DEI = 100$ as the threshold after multiple experiments and visual analysis. Due to the large number of parameters in the deblurring-stage transformer models $PMR\_T$ and $PMR\_R$, independent end-to-end training will be conducted for the final deblurring stage, constrained by server resources. All experiments were conducted on two 24GB NVIDIA GeForce RTX 3090 GPUs. $PMR$ employs a stage-wise joint training strategy. First, the $DET$ module is trained for 150 iterations with a batch size of 2, using the Adam optimizer and an image size of 272x272. The initial learning rate is $2 \times 10^{-4}$, which decays to $1 \times 10^{-6}$ with a cosine annealing scheduler. Next, the output of $DET$ is fed into $MSE\_OF$, whose output is subsequently used as the input to $DEB$. The entire $PMR$ framework is then trained for 600 iterations to optimize the full restoration pipeline, using the same batch size, optimizer, and learning rate. Other models, characterized by their end-to-end and stage-wise approaches, use a batch size of 1-2 and train for 600 iterations with the same optimizer and scheduler as $PMR$. The validation of the stage-wise joint training strategy is provided in the supplemental material.


\subsection{Experimental Results}
\textbf{Table \ref{tab:1}} compares the performance of ten methods in dynamic synthetic turbulent videos. The Normal dataset \cite{TMT} contains 1,184 videos, while the High DEI dataset includes 713 videos. Both datasets cover 32 different motion scenarios, with resolutions ranging from 480p to 960p. The three $PMR$ variants excel in most evaluation metrics, particularly on the High DEI turbulent dynamic dataset, demonstrating the effectiveness of the multi-stage framework in handling complex distortions. Notably, $PMR\_T$ outperforms all existing methods, including advanced models like $TMT$, $TurbSR$, $Deturb$, $VRT$, and $Restormer$, in the High DEI dataset. Although the lightweight $PMR$ models show slight limitations in the Normal turbulent dataset compared to physically based $TMT$, they exhibit significant advantages in inference speed and robustness in High DEI scenarios. In general, most methods perform better on the High DEI dataset than on the standard dataset, especially $TMT$, $VRT$, and the three $PMR$ methods. This indicates that training on the High DEI dataset significantly enhances the models' ability to capture complex scene features, improving performance in highly dynamic turbulent environments.

\textbf{Figure \ref{fig4}} and \textbf{Figure \ref{fig5}} further analyzes the restoration of motion and static regions in individual frames of distant and close-up video sequences from the real 2K dataset \cite{OTIS}. $TurbSR$ and $TMT$ exhibit 

\begin{figure*}[h]
  \centering
  \raisebox{-0.9\height}{%
    \resizebox{1\textwidth}{6cm}{%
      \includegraphics{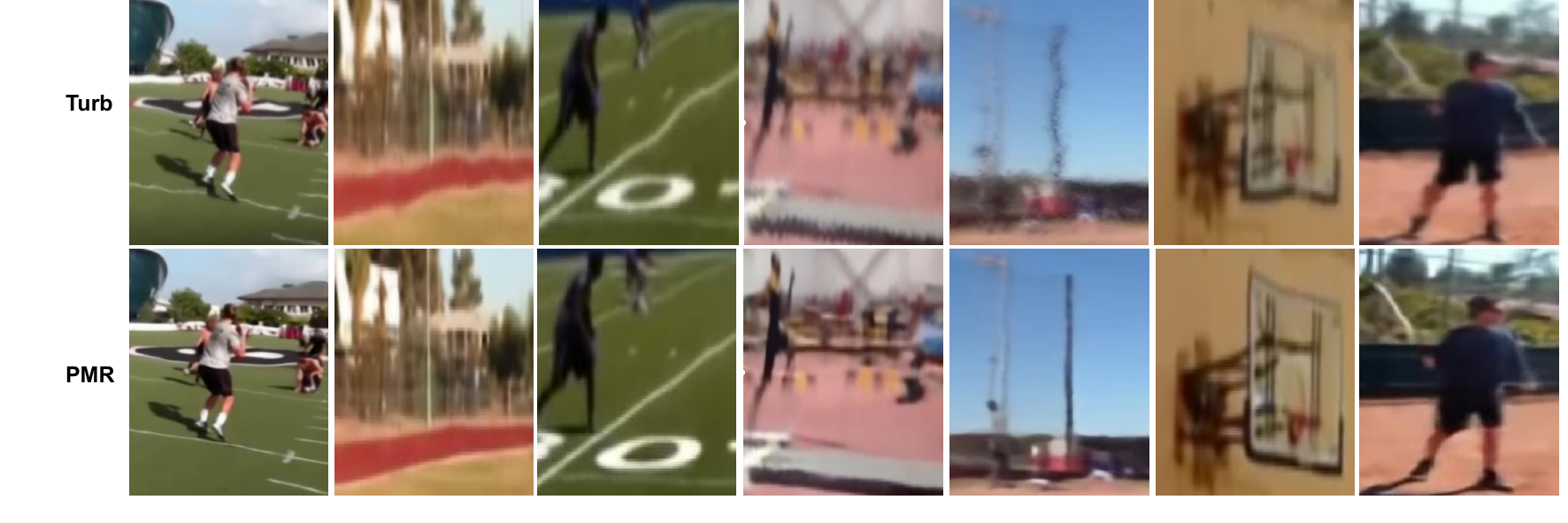}%
    }%
  }
  \caption{Visualization results of the synthetic turbulent dynamic dataset with different motion types}
  \label{fig6}
\end{figure*}

\begin{figure*}[ht]
  \centering
  \raisebox{-0.5\height}{%
    \resizebox{1\textwidth}{5.9cm}{%
      \includegraphics{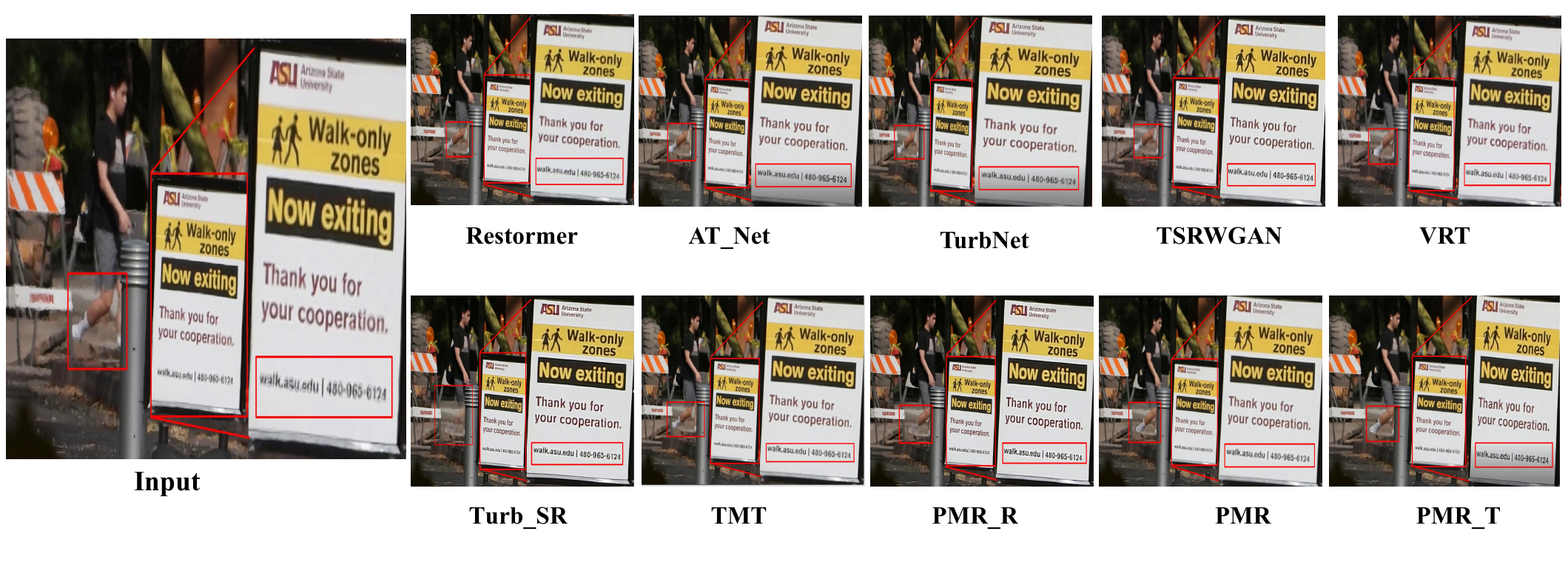}%
    }%
  }
  \caption{Comparison of text readability and detail restoration in 2K turbulent dynamic video frames from the real world.}
  \vspace{9pt} 
  \label{fig7}
\end{figure*}

\noindent motion trailing artifacts when restoring dynamic regions such as pedestrian hands. Specifically, $TurbSR$ direct segmentation recovery in heavily disturbed dynamic scenes leads to foreground misalignment and motion artifacts, while $TMT$, using average optical flow for pixel geometry correction, exacerbates edge blurring, potentially introducing artifacts in fast-moving scenes. In comparison, the three $PMR$ methods, based on physical models, preserve object edge integrity and generate clearer edge details. In dynamic scenes, $PMR$ methods accurately render moving objects, such as cars and pedestrians, significantly enhancing detail recovery and visibility, demonstrating a marked advantage. These experimental results validate the exceptional capability of the multi-stage recovery framework based on physical models in handling complex distortions in real turbulent dynamic scenes.

\textbf{Figures \ref{fig6}} present the restoration results of the proposed $PMR$ method across various synthetic turbulent motion scenarios\cite{TMT}. The results clearly show that $PMR$ achieves precise edge reconstruction and detail recovery in several key structural regions, such as baseball bats, signal poles, and running persons. The restored frames exhibit sharp object contours and rich texture details, significantly outperforming the turbulence-degraded inputs. These findings demonstrate that $PMR$ not only enhances structural fidelity and overall visibility but also maintains outstanding stability and robustness in the presence of strong turbulence and complex foreground motion.

\textbf{Figure \ref{fig7}} presents the restoration of text details in videos \cite{OTIS} by the ten methods. Single-frame models like $AT\_Net$ and $TurbNet$ introduce noticeable artifacts during detail restoration, blurring the original content. $Restormer$ and $TurbSR$ improve visual clarity in specific areas but increase information loss in regions with suboptimal restoration. $TurbSR$, in particular, produces motion artifacts around moving objects. While $TMT$ and $VRT$ show better detail recovery, they are still outperformed by $PMR\_T$ and $PMR$, which provide superior text clarity and visual quality.

\textbf{Table \ref{tab:3}} compares the proposed method with several state-of-the-art image restoration approaches in terms of inference speed, model parameters, and FLOPS across different resolutions. All methods are evaluated on an NVIDIA GeForce RTX 3090 GPU, with input frames uniformly cropped to a minimum size of 272×272 to ensure fair comparison of inference efficiency and model complexity. The results demonstrate that although our $PMR$ method involves more processing stages compared to other multi-stage approaches, its lightweight $DET$ and $DEB$ modules enable it to achieve the fastest inference speed at the 480×272 resolution. On 2K-resolution images, its inference time is only half that of $TMT$, while outperforming other multi-stage methods such as $TurbSR$ and $TMT$ in terms of parameter count and FLOPS. In addition, PMR\_R and PMR\_T, which are built upon the original Transformer-based designs of $TMT$ and $Restormer$ without further lightweight optimization, exhibit slightly higher latency than $PMR$. However, they still achieve competitive restoration performance on both synthetic and real-world datasets, while maintaining significantly fewer parameters compared to models like $Deturb$. Single-stage methods such as $VRT$, although delivering excellent restoration performance, suffer from extremely high inference time and computational cost, making them less suitable for practical deployment. Notably, while our $PMR$ model incurs additional latency compared to single-stage models such as $Uformer$ and $HistoFormer$ as image resolution increases, its stage-wise architecture offers deployment flexibility. Modules can be distributed between image acquisition and server ends, enabling resource-aware processing and maintaining acceptable response times, while achieving substantially better restoration quality than single-stage models.

\begin{table}[h]
    \caption{Comparison of Inference Speed and Model Complexity of Different Methods on Images at Various Resolutions. B, M, and S correspond to image resolutions of 1920×1088, 960×544, and 480×272, respectively; * indicates multi-stage methods. FLOPS metrics are calculated based on images with a resolution of 480×272.}
    \centering
    \setlength{\tabcolsep}{1.4pt} 
    \renewcommand\arraystretch{1.2} 
    \small 
    \begin{tabular}{c | c | c c c | c c }
    \toprule
    \multirow{2}*{\textbf{Methods}} & \multirow{2}*{\textbf{Reference}} & \multicolumn{3}{c}{\textbf{Inference speed}}$\downarrow$ & \multicolumn{2}{c}{\textbf{Complexity}}\\ \cline{3-7}
    & & $B$ & $M$ & $S$ & Param$\downarrow$ & FLOPS$\downarrow$ \\ \hline
    Restormer\cite{RESTORMER} &CVPR-2022 & 3.64s & 1.19s& 0.63s & 24.89M & 142G  \\
    Uformer\cite{uformer} &CVPR-2022 & 1.38s & 0.71s & 0.12s & 50.88M& 89G \\
    Histoformer\cite{histoformer} &ECCV-2024 & 4.13s & 2.17s & 0.46s & 16.14M & 90G  \\
    VRT\cite{10462902} &TPAMI-2024 &48.12s  &27.34s  & 6.21s & 18.33M & 7757G  \\ 
    $*$TurbSR\cite{Saha2024TurbSegResAS} & CVPR-2024 & 8.341s & 2.71s & 1.37s & 30.76M & 1723G \\
    $*$Deturb\cite{DETURB} & ACCV-2024 & 29.02s & 8.65s & 1.56s & 58.79M & 2231G  \\
    $*$TMT\cite{TMT} & TCI-2024 & 23.81s & 8.38s & 1.53S & 29.32M & 1652G  \\ \hline
    $*$PMR\_R & Ours & 15.21s     & 4.32s   & 1.52s &40.12M  & 1894G  \\
    $*$PMR & Ours & 13.88s     & 4.06s  & \textbf{0.90s} &\textbf{28.81M}  &1542G   \\
    $*$PMR\_T & Ours & 29.66s     & 8.95s   & 1.70s &34.24M  & 1762G \\ \bottomrule
    \end{tabular}
    \label{tab:3}
\end{table}

\subsection{Ablation Studies}
\paragraph{Impact of Multi-Stage Processing Order}
This study formulates the restoration of mixed distortions in turbulent dynamic videos as a sequential multi-stage process consisting of de-tilting ($DT$), segmentation enhancement ($DM$), and deblurring ($DB$). Based on the physical principles of turbulence imaging \cite{9864020}, the relationship between the clean image $\mathrm{J}(\mathrm{x})$ and turbulent image $\mathrm{I}(\mathrm{x})$ can be expressed as $\mathrm{I}(\mathrm{x}) = \mathrm{B}\left(\mathrm{T}\left(\mathrm{J}(\mathrm{x})\right)\right)$, where $T$ denotes tilt distortion and $B$ represents blur distortion. Therefore, the appropriate inversion for turbulence restoration is $\mathrm{J}(\mathrm{x}) = \mathrm{T}^{-1}\left(\mathrm{B}^{-1}\left(\mathrm{I}(\mathrm{x})\right)\right)$, indicating that applying de-tilting prior to deblurring is more effective for recovering turbulence-degraded content. Grounded in this theoretical foundation, we constrain the multi-stage pipeline to three possible processing orders. Under identical experimental conditions, we retrained the models for each order 300 iterations and evaluated their performance on 1,184 synthetic turbulent dynamic videos. As shown in Table \ref{tab:4}, the model following the $DT \rightarrow DM \rightarrow DB$ sequence achieves the best performance across all evaluation metrics, demonstrating the superiority of this processing order in the task of turbulent dynamic video restoration.

\begin{table}[H]
    \caption{Performance of PMR methods with different multi-Stage processing sequences.}
    \centering
    \renewcommand\arraystretch{1.1} 
    \begin{tabular}{l|l|l|l}
        \hline
    Processing Sequence & PSNR & SSIM & LPIPS \\
        \hline
        $DT \rightarrow DM \rightarrow DB$ & \textbf{27.216} & \textbf{0.810} & \textbf{0.256} \\
        $DM \rightarrow DT \rightarrow DB$ & 26.868 & 0.789 & 0.263 \\
        $DT \rightarrow DB \rightarrow DM$ & 27.112 & 0.803 & 0.259 \\
        \hline
    \end{tabular}
    \label{tab:4}
\end{table}

\begin{figure}[h]
  \centering
  \raisebox{-0.5\height}{%
    \resizebox{0.5\textwidth}{5cm}{%
      \includegraphics{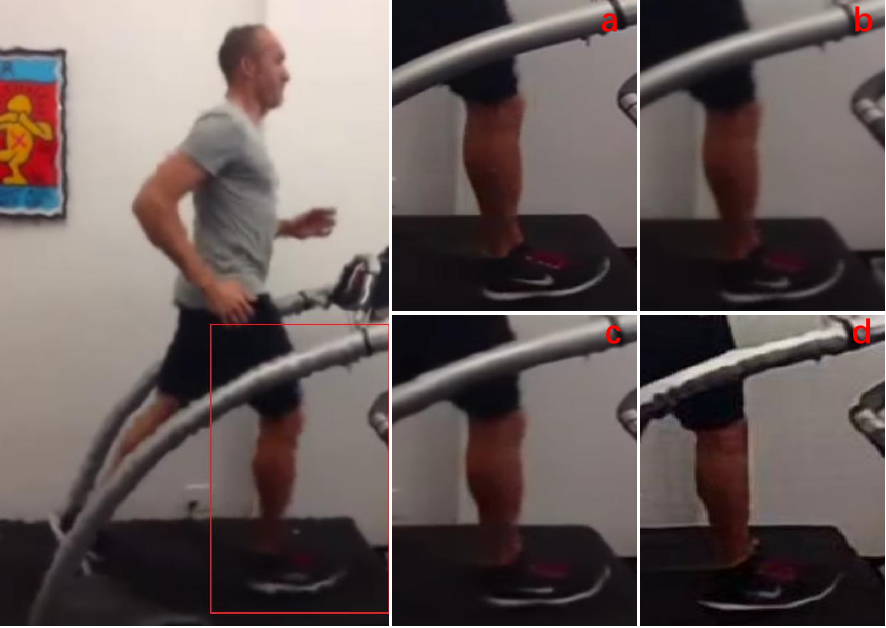}%
    }%
  }
  \caption{ Visual comparison corresponding to the ablation study in Table \ref{tab:5}.}
  \vspace{11pt} 
  \label{fig8}
\end{figure}

\paragraph{Effectiveness Verification of PMR Framework}
The $PMR$ framework decomposes the turbulence restoration task into three stages De-tilting, Motion Segmentation Enhancement, and De-blurring based on physical modeling as formulated in Equation (\ref{E1}). Each stage targets a specific type of distortion using dedicated models: $DET$, $MSE\_OF$, and $DEB$, respectively. To evaluate the contribution of each component, we conduct a series of ablation experiments by individually or jointly removing $DET$, $MSE$, or $DEB$. As shown in Table \ref{tab:5}, removing the $DET$ module leads to a more significant performance drop compared to removing MSE, highlighting its critical role in correcting global geometric distortions. When both $DET$ and $MSE$ are removed, performance degrades sharply, underscoring the importance of addressing geometric instability and motion-edge details in complex turbulent scenes. In Figure \ref{fig8}.b, removing $MSE\_OF$ results in blurred motion boundaries and loss of fine details. Figure \ref{fig8}.c shows that excluding $DET$ causes noticeable wave-like distortions in structures such as railings and legs. In Figure \ref{fig8}.d, where only the $DEB$ module is retained, severe geometric distortions and texture loss are observed, further demonstrating that deblurring alone is insufficient for effective restoration. These results confirm that each stage contributes significantly to the overall restoration quality, especially under challenging conditions involving strong turbulence and dynamic motion.

\begin{table}[H]
    \caption{Stage-wise ablation of the $PMR$ framework on the High DEI dataset, with the best results highlighted in bold.}
    \centering
    \setlength{\tabcolsep}{5pt} 
    \renewcommand\arraystretch{1.2} 
    \small 
    \begin{tabular}{l|llll}
        \hline
        $Exp.$  & $a$ & $b$ & $c$ & $d$ \\
        \hline
         $DET$      & $\checkmark$    & $\checkmark$    & $\times$ & $\times$   \\
         $MSE\_OF$  & $\checkmark$    & $\times$    & $\checkmark$ & $\times$  \\
         $DEB$      & $\checkmark$    & $\checkmark$    & $\checkmark$ &  $\checkmark$   \\ \hline
         PSNR    & \textbf{27.834 }    & 27.154  & 26.943 & 25.766  \\
         SSIM    & \textbf{0.831}     & 0.815  & 0802 & 0.764 \\
        \hline
    \end{tabular}
    \label{tab:5}
\end{table}

\paragraph{Selection of De-Blurring Models in the Multi-Stage Framework}
This study proposes three methods: $PMR\_R$, $PMR\_T$, and $PMR$, utilizing $Restormer$, $TMT$, and $DEB$ models respectively in the deblurring stage. In the synthetic High $DEI$ dataset, $PMR\_T$ shows significant advantages, as shown in Table \ref{tab:1}. In the real-world restoration effects presented in Figures \ref{fig3} and \ref{fig4}, although $PMR$ and $PMR\_R$ slightly underperform compared to PMR\_T in synthetic data, the three methods outperform other comparative methods in real-world datasets, with PMR and PMR\_T yielding the best results. Furthermore, as presented in Table \ref{tab：6}, the $DEB$ model used in $PMR$ has less than half the parameters of the $Restormer$ model in $PMR\_R$, and approximately three-quarters that of the $TMT$ model. According to Table \ref{tab:3}, $PMR$ consistently achieves the fastest inference speed across all resolutions compared to $PMR\_R$ and $PMR\_T$, highlighting its strong real-time capability. In contrast, $PMR\_T$ suffers from longer processing time and larger model size, making it less suitable for real-time deployment. $PMR\_R$ offers a more balanced speed but may introduce over-sharpening artifacts in edge details and line structures. Ultimately, $PMR$ shows the best trade-off among inference speed, model complexity, and restoration performance, and demonstrates excellent generalization ability in real-world scenarios.

\begin{table}[H]
    \caption{Comparison of transformer model parameters for deblurring tasks among $PMR$, $PMR\_R$, and $PMR\_T$.}
    \centering
    \setlength{\tabcolsep}{5pt} 
    \renewcommand\arraystretch{1.2} 
    \small 
    \begin{tabular}{l|l|l|l}
        \hline
        Methods  & PMR$\_$DEB & Restormer & TMT$\_$Transformer \\
        \hline
        Param     & \textbf{13.48M}     & 24.89M  & 19.16M  \\
        \hline
    \end{tabular}
    \label{tab：6}
\end{table}

\paragraph{Ablation Study on Lightweight Module Design in DET and DEB}
As shown in Table \ref{tab:7}, replacing the DS$\_$align module with the depth-wise 3D convolution used in $TMT$\cite{TMT} leads to a slight reduction in model parameters, but results in a drop of 0.38 dB in PSNR and 0.016 in SSIM. This demonstrates the significant contribution of optical flow-based pixel alignment to image quality improvement. Similarly, replacing the Part conv3D with a depth-wise convolution yields minimal impact on image quality metrics but increases the number of parameters, indicating the inefficiency of full-channel convolution. 
Furthermore, replacing Wave$\_$conv3D with the deformable 3D convolution from DeTurb \cite{DETURB} and substituting the STC$\_$transformer with the MDTA+GDFN structure from Restormer \cite{RESTORMER} causes only marginal performance degradation but increases the parameter count by 3.32M and 6.15M, respectively.
These results confirm that the lightweight structures in the $DET$ and $DEB$ modules are both parameter-efficient and well-suited to the spatiotemporal distortion characteristics of turbulent scenes, achieving a favorable balance between restoration quality and computational cost.

\begin{table}[h]
    \caption{Performance comparison of module replacements in DET and DEB on the High-DEI dataset.}
    \centering
    \setlength{\tabcolsep}{4.5pt} 
    \renewcommand\arraystretch{1.1} 
    \small 
    \begin{tabular}{l|ll|l}
        \hline
        Methods  & PSNR & SSIM & Param \\
        \hline
        DS$\_$align $\rightarrow$ DW conv3D  & 27.482 &0.816  & \textbf{28.72M}  \\
        Part conv3D$\rightarrow$ DW conv3D & 27.722 & 0.828  & 30.06M  \\
        Wave$\_$conv3D $\rightarrow$ Deform3D  &27.701  & 0.821  & 32.13M  \\
        STC$\_$transformer $\rightarrow$ MDTA+GDFN   &27.732 & 0.829  & 35.16M \\
        \hline
        PMR   &\textbf{27.862} & \textbf{0.832}  & 28.81M \\
        \hline
    \end{tabular}
    \label{tab:7}
\end{table}

\paragraph{Validation of the Stage-wise Joint Training Strategy}
To effectively address the compound distortions in turbulent dynamic videos, we propose a Physical Model-Driven Multi-Stage Restoration ($PMR$) framework that decomposes the restoration process into three sequential stages: de-tilting ($DET$), motion segmentation enhancement ($MSE\_OF$), and deblurring ($DEB$). Accordingly, we adopt a stage-wise joint training strategy. Specifically, the $DET$ module is first trained in an end-to-end manner. Its output is then passed to the $MSE\_OF$ module, whose output then serves as the input to the $DEB$ module. The full $PMR$ pipeline is jointly trained for final optimization.

As shown in Table \ref{tab:8}, this stage-wise joint strategy achieves the highest PSNR and SSIM on the high DEI dataset, outperforming alternative training schemes. Table \ref{tab:8}.B, which applies direct joint training across all stages without separate initialization, suffers from unstable feature propagation and parameter updates in early training. Table \ref{tab:8}.C, where all modules are trained independently in an end-to-end manner, tends to produce feature mismatches or losses between stages, leading to suboptimal results. These findings demonstrate that the proposed stage-wise joint training strategy enables better coordination across modules, thereby enhancing overall restoration performance.

\begin{table}[H]
    \caption{Comparison of performance metrics for different training strategies on the High-DEI dataset}
    \centering
    \setlength{\tabcolsep}{3pt} 
    \renewcommand\arraystretch{1.3} 
    \small 
    \begin{tabular}{l|l|l|l}
        \hline
        Methods  & A & B & C   \\
        \hline
        DET Iteration  & 50 epoch  & $\times$  &125 epoch    \\ 
        MSE$\_$OF Iteration& $\times$ &$\times$   &$\times$      \\
        DEB Iteration  & $\times$     & $\times$  &125 epoch \\
        multi-stage joint & 200 epoch     & 250 epoch &$\times$   \\
        \hline
        PSNR & \textbf{27.486}     &27.211 &26.961 \\
        SSIM & \textbf{0.822 }   & 0.815 &0.807\\
        \hline
    \end{tabular}
    \label{tab:8}
\end{table}

\paragraph{Selection of Optical Flow Algorithms in Turbulent Scenarios}
In atmospheric turbulence scenarios, image sequences frequently suffer from severe local distortions and nonlinear illumination variations, which significantly violate the brightness constancy assumption commonly used in traditional optical flow estimation. Directly applying optical flow algorithms to such turbulent sequences can introduce substantial noise and errors in the estimated flow fields, thereby impairing subsequent motion segmentation and structural restoration.

In this study, optical flow estimation serves two core purposes: first, it is a key component in computing the Dynamic Efficiency Index ($DEI$), which quantifies the magnitude and structure of motion in dynamic regions; second, it is integrated into the Motion Segmentation Enhancement module ($MSE\_OF$) to accurately separate foreground motion from static backgrounds. To mitigate the adverse effects of turbulence on optical flow estimation, we propose an adaptive regulation mechanism based on turbulence strength \(C_n^2\). Specifically, a weighting factor $\frac{1}{1 + C_n^2}$ is introduced to attenuate the impact of highly turbulent regions on flow perception. Moreover, by applying optical flow estimation after the de-tilting stage, we significantly reduce the pixel misalignment caused by turbulence, thereby enhancing the stability and reliability of motion segmentation.

To evaluate the suitability of different optical flow methods under these conditions, we conducted experiments on the URG-T dataset\cite{URGT}. As shown in Table \ref{table2}, the GMA\cite{GMA} method achieves the highest mean IoU score of 0.628, followed by RAFT\cite{RAFT} at 0.589, which outperforms lighter models such as LFNet\cite{LFNET}. Although RAFT\cite{RAFT} is not the most accurate nor the most computationally efficient method, its simpler architecture compared to GMAGMA\cite{GMA} and its strong inference stability make it well-suited for integration into our multi-stage restoration framework. Therefore, RAFT\cite{RAFT} is adopted as the primary optical flow module in this study.

\begin{table}[H]
    \caption{Mean IoU scores for different optical flow methods}
    \centering
    \setlength{\tabcolsep}{5pt} 
    \renewcommand\arraystretch{1.5} 
    \small 
    \begin{tabular}{l|l|l|l|l|l}
        \hline
        Methods  & PWC\cite{pwc}  &LFNet \cite{LFNET} & RAFT\cite{RAFT} &FF\cite{FF}  & GMA\cite{GMA}   \\
        \hline
        IoU     & 0.546   & 0.544 & 0.589  &0.622 & \textbf{0.628} \\
        \hline
    \end{tabular}
    \label{table2}
\end{table}

\section{Conclusion}
This paper proposes a multi-stage video restoration framework based on the physical model of turbulence dynamics, aimed at alleviating geometric distortions and blurring effects caused by atmospheric turbulence in long-term dynamic scenes.
The proposed $PMR$ framework achieves efficient generalization and high-quality restoration through a three-stage process: de-tilting, motion segmentation enhancement, and deblurring. Experimental results show that the proposed method performs well on the High DEI synthetic dataset, and also excels in real-world environments with high turbulence and complex dynamics, restoring details more accurately and improving visual dynamics. In general, the proposed method provides an effective solution for dynamic turbulence video restoration with significant practical application potential. Future work will further optimize the framework to address more complex scenarios with higher real-time processing demands.







\bibliography{mybibfile}

\end{document}